\pdfoutput=1
\documentclass[11pt]{article}

\usepackage[final]{acl}

\usepackage{times}
\usepackage{latexsym}
\usepackage[T1]{fontenc}
\usepackage[utf8]{inputenc}
\usepackage{microtype}
\usepackage{inconsolata}
\usepackage{hyperref}
\usepackage{url}
\usepackage{dcolumn}
\usepackage{tabularx}
\usepackage{colortbl}
\usepackage{url}
\usepackage{amssymb}
\usepackage{amsthm}
\usepackage{multirow,booktabs, hhline}
\usepackage[ruled,noend]{algorithm2e}
\usepackage{bm}
\usepackage{graphicx}
\usepackage{color}
\usepackage{enumitem}
\usepackage{makecell}
\usepackage{pifont}
\usepackage{amsmath,amsthm,amsfonts,amssymb,bm}
\usepackage{microtype}
\usepackage{xspace,mfirstuc,tabulary}
\usepackage{pifont}
\usepackage{bbm}
\usepackage{bbding}
\usepackage{subfigure}
\usepackage{makecell}
\usepackage{CJKutf8}
\usepackage{cleveref}
\usepackage{listings}
\usepackage{titlesec}
\usepackage{wrapfig}
\usepackage{lipsum}
\usepackage{hyperref}
\usepackage{tikz}
\usetikzlibrary{calc}
\usepackage{tabularray}
\usepackage{makecell}
\usepackage{multicol}
\usepackage{tcolorbox} 
\usepackage{xcolor}

\crefname{section}{§}{§§}
\Crefname{section}{§}{§§}
\definecolor{backred}{RGB}{255, 190, 190}
\definecolor{backblue}{RGB}{220, 230, 250}
\definecolor{lightergray}{RGB}{230,230,230}
\definecolor{DarkGreen}{RGB}{30,130,30}

\newcommand\ourmodel{MatchTIR\xspace}
\newcommand{\paratitle}[1]{\vspace{1.5ex}\noindent\textbf{#1}}

\newcommand{\best}{\cellcolor{backred}\textbf}
\newcommand{\high}{\cellcolor{backblue}\textbf}

\newtcbox{\hlprimarytab}{on line, rounded corners, box align=base, colback=backblue, colframe=white, size=fbox, arc=3pt, before upper=\strut, top=-2pt, bottom=-4pt, left=-2pt, right=-2pt, boxrule=0pt}
\newtcbox{\hlsecondarytab}{on line, box align=base, colback=backred, colframe=white, size=fbox, arc=3pt, before upper=\strut, top=-2pt, bottom=-4pt, left=-2pt, right=-2pt, boxrule=0pt}
\newcommand{\besttext}[1]{{\hlsecondarytab{#1}}}
\newcommand{\hightext}[1]{{\hlprimarytab{#1}}}

\newenvironment{itemize*}%
 {\leftmargini=20pt\begin{itemize}%
  \setlength{\itemsep}{3pt}%
  \setlength{\parskip}{0pt}%
  }%
 {\end{itemize}}
\newenvironment{enumerate*}%
 {\begin{enumerate}%
  \setlength{\itemsep}{0pt}%
  \setlength{\parskip}{0pt}}%
 {\end{enumerate}}
\definecolor{chart Idle}{gray}{.6}
\definecolor{chart Poor}{RGB}{242,28,28}
\definecolor{chart Ok}{RGB}{248,172,37}
\definecolor{chart Ideal}{RGB}{1,151,0}
\definecolor{chart Over}{RGB}{0,125,234}
\definecolor{Gainsboro}{rgb}{0.86, 0.86, 0.86}
\definecolor{MidnightBlue}{RGB}{43, 101, 144}
\definecolor{lightgold}{rgb}{1.0, 0.95, 0.8}
\definecolor{ablationblue}{RGB}{236, 244, 255}
\newdimen\tempdim

\newcommand*{\ChartBox}[3]{%
  \begingroup
    \settoheight{\tempdim}{L}%
    \edef\tempheight{\the\tempdim}%
    \settodepth{\tempdim}{g}%
    \edef\tempdepth{\the\tempdim}%
    \tikz[
      baseline=0pt,
      inner sep=0pt,
    ]
    \node[
      fill={#3!#2},
      rounded corners=1pt,
      anchor=base,
    ]{%
      \vphantom{g\"A}%
      \pgfmathsetlength{\tempdim}{#1}%
      \kern\tempdim\relax
    };%
  \endgroup
}

\title{MatchTIR: Fine-Grained Supervision for \\ Tool-Integrated Reasoning via Bipartite Matching}

\author{
    Changle Qu\textsuperscript{1},
    Sunhao Dai\textsuperscript{1},
    Hengyi Cai\textsuperscript{2},
    Jun Xu\textsuperscript{1},
    Shuaiqiang Wang\textsuperscript{2},
    Dawei Yin\textsuperscript{2} \\
    \textsuperscript{1}Gaoling School of Artificial Intelligence, Renmin University of China; 
    \textsuperscript{2}Baidu Inc. \\
    \texttt{\{changlequ,sunhaodai,junxu\}@ruc.edu.cn},
    \texttt{caihengyi@ict.ac.cn},\\
    \texttt{wangshuaiqiang@baidu.com}, \texttt{yindawei@acm.org}
}

\begin{document}
\maketitle
\begin{abstract}
Tool-Integrated Reasoning (TIR) empowers large language models (LLMs) to tackle complex tasks by interleaving reasoning steps with external tool interactions.
However, existing reinforcement learning methods typically rely on outcome- or trajectory-level rewards, assigning uniform advantages to all steps within a trajectory. 
This coarse-grained credit assignment fails to distinguish effective tool calls from redundant or erroneous ones, particularly in long-horizon multi-turn scenarios.
To address this, we propose MatchTIR, a framework that introduces fine-grained supervision via bipartite matching-based turn-level reward assignment and dual-level advantage estimation.
Specifically, we formulate credit assignment as a bipartite matching problem between predicted and ground-truth traces, utilizing two assignment strategies to derive dense turn-level rewards.
Furthermore, to balance local step precision with global task success, we introduce a dual-level advantage estimation scheme that integrates turn-level and trajectory-level signals, assigning distinct advantage values to individual interaction turns.
Extensive experiments on three benchmarks demonstrate the superiority of \ourmodel. 
Notably, our 4B model surpasses the majority of 8B competitors, particularly in long-horizon and multi-turn tasks.
Our codes are available at~\url{https://github.com/quchangle1/MatchTIR}.
\end{abstract}

\section{Introduction}
\label{introsection}

\begin{figure}[t]
\centering
	\includegraphics[width=\linewidth]{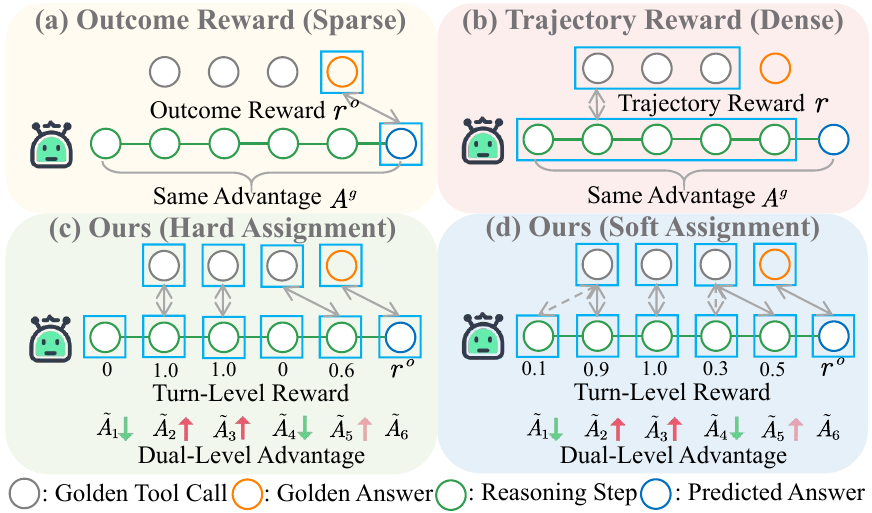}
\caption{Comparison of reward and advantage assignment strategies for multi-turn TIR. (a) and (b) show traditional methods where all reasoning steps share the same reward and advantage. (c) and (d) illustrate our \ourmodel framework using hard and soft assignment to derive turn-level rewards and distinct advantages.}
        \vspace{-0.4cm}
        \label{fig:intro_a}
        \vspace{-0.4cm}
\end{figure}

Tool-Integrated Reasoning (TIR) has established itself as a critical paradigm for enhancing the capabilities of large language models~(LLMs) by enabling them to interact with external tools during the reasoning process~\cite{gou2023tora}. 
By leveraging external tools, TIR allows LLMs to overcome limitations of static parametric knowledge~\cite{jin2025search,zheng2025deepresearcher}, perform precise computations~\cite{wang2024mathcoder,das2024mathsensei}, and engage dynamically with external environments~\cite{qin2023toolllm,qu2024colt,qu2025from}. 
Through multi-turn interactions, agents interleave reasoning, tool execution, and feedback acquisition, refining their trajectory toward solving complex real-world tasks.

To equip agents with such sophisticated tool-integrated reasoning capabilities, recent research has largely adopted reinforcement learning with verifiable rewards (RLVR) methods \cite{chang2025thor,jiang2025verltool,lin2025understanding}, such as Group Relative Policy Optimization (GRPO) \cite{shao2024deepseekmath}.
While early RLVR formulations primarily relied on sparse outcome-based rewards~\cite{li2025torl}, subsequent works have incorporated trajectory-level signals~\cite{qian2025toolrl,zeng2025tool} to provide denser supervision.
However, as shown in Figure~\ref{fig:intro_a}, even these trajectory-level formulations typically assign a uniform advantage value to every turn within a trajectory. 
This ``one-size-fits-all'' credit assignment fails to distinguish critical reasoning steps from redundant or erroneous tool calls, leading to inefficient optimization and consequently hindering the model from learning precise and efficient tool-use strategies.

Several studies have attempted to introduce fine-grained rewards via intrinsic signals or external reward models~\cite{wang2025information,zhang2025criticsearch}.
However, external reward models are inherently susceptible to bias and hallucination, while Monte Carlo estimation~\cite{dong2025agentic} incurs prohibitive computational costs and high variance, especially in long-horizon tasks.
Moreover, many of these efforts are confined to restricted tools such as search engines, where intermediate steps are characterized by high semantic variance due to the fact that multiple different queries may be equally valid, making it difficult to establish a unique ground truth for reliable step-level supervision.
In contrast, general TIR scenarios naturally expose structured and verifiable signals, such as tool names, parameter names, and parameter contents, which enable explicit evaluation of the correctness of the tool use at each turn.
This motivates our approach to formulate turn-level reward assignment as a bipartite matching problem between predicted and ground-truth tool interactions.

In this paper, we propose \ourmodel, a framework designed to assign distinct, precise advantages to individual turns within a reasoning trajectory.
We formulate turn-level credit assignment as a bipartite matching problem, constructing a weighted bipartite graph based on similarity scores across tool names, parameter names, and parameter contents to align predicted calls with ground-truth references.
Based on this formulation, we introduce both hard and soft credit assignment strategies to derive dense turn-level rewards:
the hard variant enforces ``one-to-one'' matching, while the soft variant allows ``one-to-many'' alignment.
Both strategies produce dense and precise turn-level rewards, enabling effective supervision of intermediate tool interactions.
To ensure the model optimizes for both local accuracy and global success, \ourmodel synthesizes turn-level rewards with final outcome-based rewards. 
Furthermore, we introduce a dual-level advantage estimation mechanism that harmonizes local turn-level rewards with global trajectory-level outcomes. 
Specifically, the trajectory-level advantage aggregates all rewards across the entire path to evaluate global quality, while the turn-level advantage captures the $t$-th turn's specific contribution via discounted accumulation of all subsequent rewards.
The policy is finally optimized using the GRPO objective with our integrated dual-level advantages.

Extensive experiments on both in-domain and out-of-domain benchmarks demonstrate the effectiveness and robustness of \ourmodel.
In summary, our main contributions are as follows:

$\bullet$~We identify the uniform credit assignment problem in existing TIR methods as a key bottleneck preventing efficient optimization.

$\bullet$~We propose \ourmodel, which formulates turn-level credit assignment as a bipartite matching problem and introduces hard and soft matching strategies to provide dense, turn-level supervision.

$\bullet$~Extensive experiments on three benchmarks demonstrate the effectiveness of \ourmodel, with particularly strong gains on long-horizon scenarios, validating its robustness and generalizability.

\section{Preliminaries}
In this section, we first present the task formulation of TIR, and then introduce the reinforcement learning framework adopted in this work.

\subsection{Task Formulation}
Formally, given a user query $q$ and a tool set $\mathcal{T} = \{t_1, t_2, \ldots, t_{n}\}$ consisting of $n$ available tools, the goal of the agent is to generate an interaction trajectory $\tau = \{s_1, s_2, \ldots, s_T\}$ to solve $q$, where $T$ denotes the number of interaction turns.
Specifically, each interaction turn $s_i = (n_i,c_i,o_i)$ is represented as a triplet, where $n_i$ is the natural language reasoning in $i$-th turn, $c_i$ denotes the set of invoked tools from $\mathcal{T}$ along with their specific parameter names and parameter contents, and $o_i$ is the observation returned by the environment. 
The trajectory terminates when the agent generates a final answer response (where $c_T = \emptyset$ and $o_T = \emptyset$) or reaches the pre-defined maximum turn limit $L$.

\paratitle{Goal.} As discussed in \S\ref{introsection}, TIR is inherently a long-horizon, multi-turn process. 
However, existing outcome- or trajectory-level reward formulations assign uniform credit across an entire trajectory, failing to distinguish between effective and unnecessary tool invocations.
This results in inefficient credit assignment and suboptimal tool-use behaviors.
Therefore, our goal is to assign precise turn-level rewards, enabling fine-grained credit assignment that incentivizes effective tool usage contributing to the final solution while penalizing redundant or erroneous actions at each turn.

\begin{figure*}[t]
\centering
	\includegraphics[width=\linewidth]{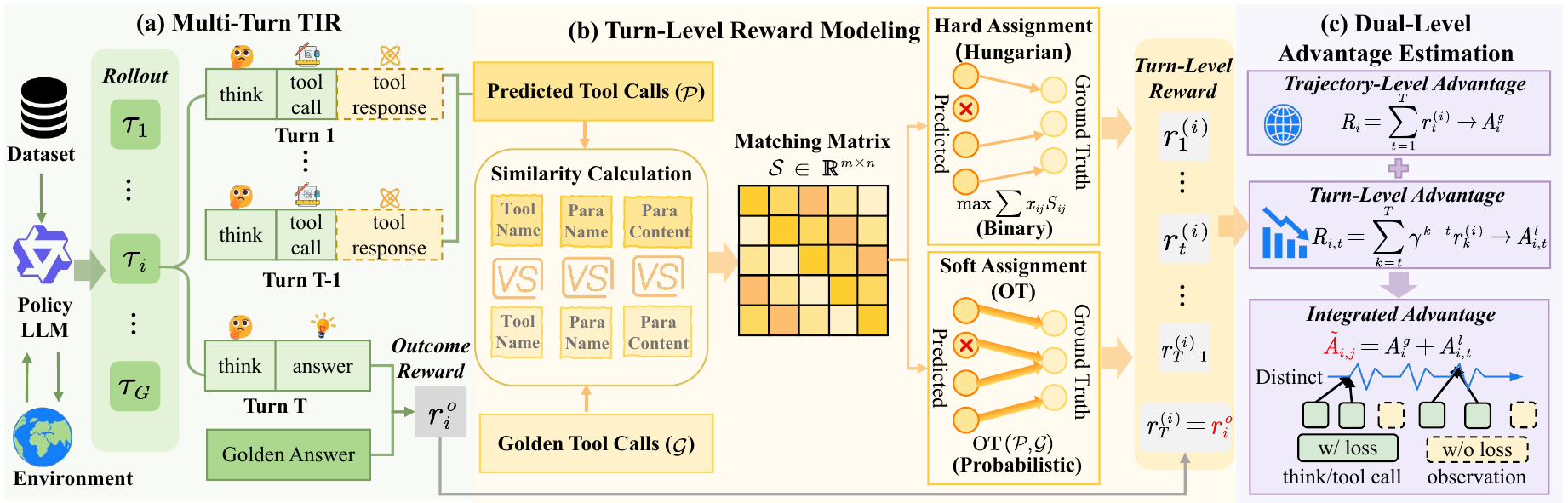}
    % \vspace{-0.2cm}
\caption{The illustration of our proposed \ourmodel. (a) illustrates the process of multi-turn TIR, where the policy LLM interacts with external tools over multiple turns and receives an outcome reward. (b) shows turn-level reward modeling by matching predicted and golden tool calls using hard or soft assignment strategies. (c) depicts the dual-level advantage estimation that integrates trajectory-level and turn-level signals for policy optimization.}
        \label{fig:method}
        \vspace{-0.6cm}
\end{figure*}

\subsection{Agentic Reinforcement Learning}
To optimize the agent policy $\pi_{\theta}$, we adopt Group Relative Policy Optimization (GRPO)~\cite{shao2024deepseekmath}, a specialized reinforcement learning algorithm. 
Distinct from conventional methods, GRPO obviates the requirement for a learned value function by estimating advantages based on the relative performance within a group of $G$ rollouts $\{\tau_i\}_{i=1}^{G}$ generated from the same prompt.
The GRPO training objective can be formulated as

\begin{small}
\begin{equation*}
    \begin{split}
        &\mathcal{J}_{\text{GRPO}}(\theta)=\mathbb{E}_{q\sim D,\{\tau_i\}\sim\pi_{\theta_\text{old}}(\cdot|q)}\frac{1}{G}\sum_{i=1}^{G}\frac{1}{|\tau_i|}\sum_{j=1}^{|\tau_i|}\bigg[\min(\\
        &w_{i,j}\hat{A}_{i,j},\text{clip}(w_{i,j},
        1-\epsilon,1+\epsilon)\hat{A}_{i,j})-\beta \mathbb{D}_\text{KL}(\pi_{\theta}|\pi_\text{ref})\bigg],
    \end{split}
\label{eq:grpo}
\end{equation*}
\end{small}
where $\pi_{old}$ is the old policy, $w_{i,j} = \frac{\pi_{\theta}(\tau_{i,j}|q, \tau_{i,<j})}{\pi_{\theta_{\text{old}}}(\tau_{i,j}|q, \tau_{i,<j})}$ is the importance sampling ratio, $\hat{A}_{i,j}$ is the normalized advantage of the $j$-th token in the $i$-th rollout.
The hyper-parameters $\epsilon$ and $\beta$ denote the clipping threshold and the coefficient for KL divergence, respectively, where the latter regularizes the policy to remain close to the reference model $\pi_{\text{ref}}$.
During training, tokens corresponding to tool responses are masked out when computing the objective, since these tokens are generated by the external environment rather than being controlled by the agent.

\section{Our Approach: \ourmodel}

In this section, we will provide a detailed introduction to our proposed \ourmodel. 
The overall framework is illustrated in Figure~\ref{fig:method}.

\subsection{Reward Modeling}
As shown in Figure~\ref{fig:method}(a), TIR is inherently a multi-turn process, where an agent interleaves reasoning with multiple tool invocations before arriving at a final answer.
To optimize the agent's policy, it is crucial to determine whether each predicted tool call contributes to the reasoning goal.
However, simply comparing individual calls is insufficient, as the agent may generate redundant tool invocations.
To address this, we reformulate turn-level credit assignment for TIR as a bipartite matching problem between two sets.
Specifically, given a rollout trajectory, we extract a set of $m$ predicted tool calls $\mathcal{P} = \{p_1, \dots, p_m\}$ and identify $n$ ground-truth tool calls $\mathcal{G} = \{g_1, \dots, g_n\}$. 
Our goal is to derive precise rewards by finding an optimal alignment between $\mathcal{P}$ and $\mathcal{G}$, thereby penalizing uninformative or repetitive actions while rewarding those that align with the expert reasoning path.

\paratitle{Matching Matrix Construction.} We first construct a matching matrix $\mathcal{S} \in \mathbb{R}^{m \times n}$, where each entry $S_{ij}$ represents the matching score between a predicted tool $p_i$ and a ground-truth tool $g_j$.
Inspired by the previous work~\cite{qian2025toolrl}, the matching score $S_{ij}$ includes three components:

$\bullet $ Tool Name Matching. We first check the consistency of the tool name. Let $p_i^{tn}$ and $g_j^{tn}$ be the name of the predicted and golden tool, respectively:
\begin{equation}
    S_{tn} = \mathbb{I}(p_i^{tn} = g_{j}^{tn})\in \{0,1\},
\end{equation}
where $\mathbb{I}(\cdot)$ is an indicator function that returns 1 if the predicted tool name matches the ground-truth tool name, and 0 otherwise.

$\bullet $ Parameter Name Matching. If the tool name matches, we calculate Jaccard similarity of parameter name sets $N_{p_i}$ and $N_{g_j}$ to measure the overlap:
\begin{equation}
    S_{pn} = \frac{|N_{p_i} \cap N_{g_j}|}{|N_{p_i} \cup N_{g_j}|} \in [0,1],
\end{equation}

$\bullet $ Parameter Content Matching. We further assess the correctness of the values provided for each ground-truth parameter:
\begin{equation}
    S_{pc} =\sum_{k\in N_{g_j}} \mathbb{I}(p_i[k] = g_j[k]) \in [0,|N_{g_j}|],
\end{equation}
where $p_i[k]$ and $g_j[k]$ denote the parameter content associated with the parameter name $k$.

The final similarity score $S_{ij}$ is normalized by the maximum possible score to ensure $S_{ij} \in [0, 1]$:
\begin{equation}
S_{ij} = S_{tn} \cdot \frac{S_{tn} + S_{pn} + S_{pc}}{1+1 + |N_{g_j}|}
\end{equation}

While the matrix $\mathcal{S}$ provides pairwise alignment scores, directly using these scores as rewards can lead to reward hacking, such as repeatedly invoking similar tools to accumulate high similarity scores. 
To mitigate this, we introduce two assignment strategies to convert $\mathcal{S}$ into turn-level rewards.

\paratitle{Hard Credit Assignment.}
The hard credit assignment strategy enforces a strict ``one-to-one'' alignment between predicted and ground-truth tool calls, preventing duplicated credit for the same ground-truth action.
We formulate this process as a maximum weight bipartite matching problem and solve it using the Hungarian (Kuhn–Munkres) algorithm~\cite{kuhn1956variants}.
Formally, let $x_{ij} \in \{0,1\}$ be a binary assignment variable that $x_{ij} =1$ implies that the predicted tool $p_i$ is matched with the ground-truth tool $g_j$. 
The objective is to maximize the total matching score $\sum_{i,j}x_{i,j}S_{i,j}$, subject to the constraint that each predicted and ground-truth tool call is matched at most once.
Based on the resulting matching, the reward for each predicted tool call $p_i$ is defined as
\begin{equation}
r_{p_i} = \begin{cases}
S_{ij}, & \text{if } \exists j, x_{ij} = 1 \\
-\lambda, & \text{otherwise} 
\end{cases}
\end{equation}
where $\lambda$ is a fixed penalty (or zero) assigned to erroneous or redundant tool calls that fail to match any ground-truth tool call.

\paratitle{Soft Credit Assignment.}
To facilitate a smoother optimization landscape and provide more nuanced feedback, we propose a soft relaxation of the assignment problem based on Optimal Transport (OT)~\cite{cuturi2013sinkhorn}.
Instead of enforcing a hard binary choice, this method treats the predicted and ground-truth tool sequences as discrete probability distributions and optimizes the cost of transforming one into the other.
The matching matrix $\mathcal{S}$ is transformed into a cost matrix $\mathcal{C}$ via a decreasing transformation (e.g., $C_{ij} = - S_{ij}$), ensuring that high matching scores translate to minimal transportation costs.
We then compute an optimal transport plan $\mathcal{Z}$ that minimizes the total cost:
\begin{equation} 
\min_{\mathcal{Z}} \sum_{i,j} Z_{ij} C_{ij}, \;\;\; \text{s.t. } \mathcal{Z}\mathbf{1}_n = \mathbf{a}, \mathcal{Z}^\top\mathbf{1}_m = \mathbf{b} 
\end{equation}
% \vspace{-1em}
where $\mathbf{a} \in \mathbb{R}^m$ and $\mathbf{b} \in \mathbb{R}^n$ are marginal distributions representing the importance weights of predicted and ground-truth tool calls, respectively.

The resulting transport plan $\mathcal{Z}$ effectively functions as a probabilistic mapping between predicted and ground-truth interactions.
Given the resulting transport plan, the reward $r_{p_i}$ for each predicted tool call is then calculated as the weighted sum of its similarities across the ground-truth set:
\begin{equation}
    r_{pi} = \sum_{j}Z_{ij} \cdot S_{ij}.
\end{equation}

Unlike the hard version, soft assignment allows a single ground-truth tool call reward to be distributed across multiple predicted turns. 
This mechanism ensures that the turn with the highest alignment receives the largest share of credit, while other relevant attempts still receive partial, dense feedback to guide the policy's improvement.

\paratitle{Turn-Level Reward.}
The assignment mechanisms described above yield a specific reward $r_{p_i}$ for each individual predicted tool call.
However, within the multi-turn TIR framework, a single interaction turn may involve multiple tool calls, whereas policy optimization is typically performed at the turn level. 
Therefore, it is necessary to aggregate tool call rewards into a unified turn-level signal.
To this end, we compute the turn-level reward by averaging the rewards of predicted tool calls within the same turn:
\begin{equation} 
r_t = \frac{1}{|P_t|} \sum_{p_i \in P_t} r_{p_i}, 
\end{equation}
where $|P_t|$ denotes the number of predicted tool calls issued at turn $t$.
This averaging strategy normalizes the reward across turns and discourages excessive tool usage, ensuring fair comparison between turns with different numbers of tool calls.

\paratitle{Outcome-Level Reward.}
To complement the fine-grained supervision provided by turn-level rewards, we incorporate an outcome-level reward $r^o$ to evaluate the global objective fulfillment of the entire reasoning trajectory.
Specifically, we use the F1 score between the predicted answer $\hat{a}$ and the golden answer $a$ as the outcome-level reward:
\begin{equation} 
r^o = \frac{2|\hat{a}\cap a|}{|\hat{a}|+|a|} \in [0,1].
\end{equation}

This outcome-level reward serves as a global constraint that complements the turn-level supervision, encouraging the model to not only make correct tool calls at each turn but also remain aligned with successful end-task completion.

\subsection{Dual-Level Advantage Estimation}
In conventional outcome-supervised RL frameworks like GRPO, all tokens within a sampled rollout are typically assigned a uniform advantage value derived from the final outcome reward. 
While this strategy is effective for single-turn or short-horizon tasks, it is fundamentally suboptimal for TIR, where a trajectory consists of multiple interaction turns with highly heterogeneous contributions.
Assigning identical advantages to all tokens fails to distinguish between informative actions that advance task completion and uninformative or redundant tool-use behaviors, resulting in coarse credit assignment and noisy policy optimization.
To address this limitation, we propose a dual-level advantage estimation framework that evaluates model actions from two complementary perspectives:

\paratitle{Trajectory-Level Advantage Estimation.}
We first evaluate the overall quality of each rollout within a group of $G$ sampled trajectories. 
For the $i$-th rollout, we define the trajectory-level reward $R_i$ as the sum of its turn- and outcome-level rewards:
\begin{equation}
R_i = \sum_{t=1}^{T} r_{t}^{(i)},
\end{equation}
where $r_{t}^{(i)}$ is the turn-level reward at turn $t$ of the $i$-th rollout. 
For the final turn $T$, $r_{T}^{(i)}$ corresponds to the outcome-level reward $r_i^o$ .

Following the GRPO framework, the trajectory-level advantage $A_i^g$ is computed by normalizing the rewards across the group:$
A_i^g = \frac{R_i - \text{mean}(\{R_j\}_{j=1}^{G})}{\text{std}(\{R_j\}_{j=1}^{G})}.$
This advantage reflects the relative performance of the entire reasoning path compared to others generated for the same prompt, providing a global supervisory signal for policy optimization.

\paratitle{Turn-Level Advantage Estimation.}
While trajectory-level advantage provides a global performance signal, it assigns identical supervision to all interaction turns within the trajectory.
To further differentiate the contributions of individual turns, we introduce local turn-level advantage estimation, which provides fine-grained supervision aligned with turn-level rewards.
Specifically, to capture the impact of current decisions on future turns, we first compute a discounted cumulative reward $R_{i,j}$ for each turn $t$ in the $i$-th rollout:
\begin{equation}
R_{i,t} = \sum_{k=t}^{T} \gamma^{k-t} r_{k}^{(i)},
\end{equation}
where $\gamma$ is the discount factor. 
This formulation encourages the model to account for the long-term consequences of its tool-use behaviors.

To provide fine-grained supervision, we then compute the local turn-level advantage $A_{i,t}^l$ by comparing the cumulative reward  against its group-level baseline: $A_{i,t}^l = \frac{R_{i,t} - \text{mean}(\{R_{j,t}\}_{j=1}^{G})}{\text{std}(\{R_{j,t}\}_{j=1}^{G})}.$
This term captures fine-grained differences across interaction turns.
This dual-view advantage estimation enables more precise credit assignment for TIR.

\subsection{Policy Optimization}
\paratitle{Advantage Integration.} 
To jointly leverage the global task completion signal and local turn-level supervision, we integrate the trajectory-level and turn-level advantages into a unified supervisory signal.
For a token at position $j$ belonging to turn $t$ within the $i$-th rollout, the integrated advantage $\tilde{A}_{i,j}$ is defined as $\textcolor{red}{\tilde{A}_{i,j}} = A_i^g + A_{i,t}^l.$ 

\paratitle{Objective Function.}
Using the integrated advantage, we optimize the policy $\pi_\theta$ under the GRPO framework by replacing the advantage with $\tilde{A}_{i,j}$:

\begin{small}
\begin{equation*}
    \begin{split}
        \mathcal{J}&(\theta)=\mathbb{E}_{q\sim D,\{\tau_i\}\sim\pi_{\theta_\text{old}}(\cdot|q)}\frac{1}{G}\sum_{i=1}^{G}\frac{1}{|\tau_i|}\sum_{j=1}^{|\tau_i|}\bigg[\min(w_{i,j}\\
        &\textcolor{red}{\tilde{A}_{i,j}},\text{clip}(w_{i,j},
        1-\epsilon,1+\epsilon)\textcolor{red}{\tilde{A}_{i,j}})-\beta \mathbb{D}_\text{KL}(\pi_{\theta}|\pi_\text{ref})\bigg].
    \end{split}
\label{eq:our_grpo}
\end{equation*}
\end{small}
\section{Experiments}
In this section, we describe the experimental setups and present a comprehensive evaluation, with additional results provided in appendix~\ref{sec:appendix_experiments}.

\begin{table*}[t]
	\centering
        \resizebox{1.0\textwidth}{!}{
	\begin{tabular}{l|cccc|ccccccc|c|c}
        \toprule
        \multirow{3}{*}{\textbf{Methods}}
        & \multicolumn{4}{c|}{\textbf{FTRL}}
        & \multicolumn{7}{c|}{\textbf{BFCL}}
        & \multicolumn{1}{c|}{\textbf{ToolHop}}
        & \multirow{3}{*}{\textbf{Avg.}} \\
        \cmidrule(lr){2-5}
        \cmidrule(lr){6-12}
        \cmidrule(lr){13-13}
        
        & \multirow{2}{*}{Solve-P}
        & \multirow{2}{*}{Solve-R}
        & \multirow{2}{*}{Solve-F1}
        & \multirow{2}{*}{Avg.}
        & \multicolumn{4}{c}{Multi-Turn}
        & \multicolumn{2}{c}{Agentic}
        & \multirow{2}{*}{Avg.}  & \multirow{2}{*}{AC} \\
        \cmidrule(lr){6-9}
        \cmidrule(lr){10-11}
        
        & & & &
        & Base & MF & MP & LC
        & Search & Memory &
        & 
        &  \\
        \hline
        \rowcolor{Gainsboro} \multicolumn{14}{c}{\textit{\textbf{Qwen3-4B}}} \\
        % \multicolumn{14}{l}{\textbf{Qwen3-4B}} \\
        \midrule
        Vanilla
        & 30.78 & 29.65 & 25.85 & 28.76
        & 41.50 & 31.00 & 26.50 & 27.50
        & 9.00 & 19.57 & 22.96
        & 31.63
        & 27.78 \\
        
        GRPO
        & 31.13 & 32.83 & 30.67 & 31.54
        & 45.00 & 37.50 & 26.50 & 29.50
        & 9.50 & 21.94 & 25.17
        & 37.25
        & 31.32 \\
        
        FTRL-S
        & 34.10 & 34.07 & 31.54 & 33.24
        & 43.00 & 35.50 & 31.00 & 28.00
        & 6.50 & \best{23.44} & 24.67
        & 38.63
        & 32.18 \\
        
        FTRL-M
        & \best{34.47} & 35.44 & 33.50 & 34.47
        & 44.50 & 41.00 & \best{32.50} & 29.50
        & 6.50 & 22.80 & 25.76
        & 41.24
        & 33.82 \\
        
        ToolRL-S
        & 28.26 & 28.32 & 23.78 & 26.79
        & 32.50 & 31.00 & 22.50 & 20.00
        & 7.50 & 19.14 & 19.91
        & 30.28
        & 25.66 \\
        
        ToolRL-M
        & 30.90 & 36.26 & 32.28 & 33.15
        & 50.00 & 37.50 & 29.00 & 30.00
        & 10.50 & 22.80 & 26.64
        & 35.68
        & 31.82 \\
        \midrule
        \textbf{Ours (OT)}
        & 31.79 & 37.52 & 32.60 & 33.97
        & 50.00 & 40.50 & 26.50 & 35.00
        & 12.00 & 21.72 & 27.43
        & 41.95
        & 34.45 \\
        
        \textbf{Ours (KM)}
        & 32.39 & \best{39.70} & \best{34.21} & \best{35.43}
        & \best{50.50} & \best{47.00} & 28.50 & \best{36.50}
        & \best{17.50} & 22.15 & \best{30.23}
        & \best{42.55}
        & \best{36.07} \\
        
        \hline
        \rowcolor{Gainsboro} \multicolumn{14}{c}{\textit{\textbf{Qwen3-8B}}} \\
        \midrule
        
        Vanilla
        & 28.08 & 36.55 & 29.74 & 31.46
        & 47.50 & 46.00 & 37.50 & 34.50
        & 10.00 & 20.43 & 28.30
        & 42.21
        & 33.99 \\
        
        GRPO
        & 31.59 & 39.75 & 32.54 & 34.63
        & 52.50 & 45.50 & 34.50 & 36.50
        & 14.50 & 18.92 & 29.48
        & 40.64
        & 34.92 \\
        
        FTRL-S
        & 32.32 & 38.87 & 32.85 & 34.68
        & 51.50 & 45.00 & 35.50 & 34.00
        & 12.50 & 17.42 & 28.23
        & 36.72
        & 33.21 \\
        
        FTRL-M
        & 33.28 & 40.60 & 34.20 & 36.03
        & 50.50 & 51.00 & 35.00 & 33.50
        & 15.00 & 21.94 & 30.49
        & 43.32
        & 36.61 \\
        
        ToolRL-S
        & 25.57 & 35.31 & 26.72 & 29.20
        & 41.00 & 39.50 & 31.50 & 25.00
        & 7.50 & 17.63 & 23.41
        & 32.93
        & 28.51 \\
        
        ToolRL-M
        & 30.64 & 40.07 & 32.51 & 34.41
        & 50.00 & 47.00 & \high{39.50} & 32.00
        & 15.50 & 16.77 & 29.13
        & 42.55
        & 35.36 \\
        \midrule
        \textbf{Ours (OT)}
        & 33.61 & 42.56 & 33.61 & 36.59
        & 55.50 & \high{52.00} & 38.50 & 36.00
        & \high{18.00} & 23.87 & 33.22
        & 45.80
        & 38.54 \\
        
        \textbf{Ours (KM)}
        & \high{36.33} & \high{44.18} & \high{37.33} & \high{39.28}
        & \high{60.00} & 49.00 & 39.00 & \high{40.50}
        & 17.00 & \high{26.02} & \high{34.32}
        & \high{46.16}
        & \high{39.92} \\
        
        \bottomrule
        \end{tabular}
	}
    \vspace{-0.2cm}
	\caption{Performance comparison between \ourmodel and the baselines on three benchmarks with two LLMs. For our method, KM and OT denote the hard credit assignment and the soft credit assignment, respectively. For BFCL, MF, MP, and LC correspond to the Miss Function, Miss Parameter, and Long Context subsets, respectively. We highlight the best performance of Qwen3-4B models and Qwen3-8B models with the \besttext{red} and \hightext{blue}, respectively.}
	\label{tab:main}
    \vspace{-0.4cm}
\end{table*}

\subsection{Experimental Setups}
\paratitle{Training Dataset.}
We train the model on the FTRL dataset~\cite{ye2025feedback}, which comprises over 2,000 automatically constructed tool-use environments with verifiable feedback, covering single-hop, multi-hop, and parallel reasoning scenarios.

\paratitle{Evaluation Benchmarks.}
We evaluate \ourmodel on three benchmarks: 
FTRL is used for in-domain evaluation, while BFCL~\cite{patil2025bfcl} and ToolHop~\cite{ye2025toolhop} are adopted for out-of-domain evaluation to assess generalization.
Details of these benchmarks are shown in Appendix~\ref{sec:appendix_benchmark}.

\paratitle{Baselines.}
To evaluate the effectiveness of our proposed \ourmodel, we compare it against comprehensive baselines:
(1) Vanilla: We evaluate the vanilla Qwen3~\cite{yang2025qwen3} models without reinforcement learning.
(2) GRPO~\cite{shao2024deepseekmath}: The standard GRPO framework that optimizes the policy solely based on outcome rewards.
(3) ToolRL~\cite{qian2025toolrl}: ToolRL assigns rewards based on the correctness of tool names, parameter names, and parameter content. 
While originally designed for single-turn tool use, we extend it to the multi-turn interaction setting by applying its scoring mechanism at the trajectory level.
The single-turn and multi-turn variants are denoted as ToolRL-S and ToolRL-M, respectively.
(4) FTRL~\cite{ye2025feedback}: FTRL computes rewards using the ratio of successful tool calls to total attempts.
Similar to ToolRL, we include both the single-turn baseline (FTRL-S) and its adaptation to the multi-turn setting (FTRL-M).

\paratitle{Implementation Details.}
We adopt Qwen3-4B and Qwen3-8B as backbone models and optimize them using the GRPO objective implemented in the verl framework~\cite{sheng2025hybridflow}.
All experiments are conducted under the Zero-RL setting, with a batch size of 256 and 16 rollouts sampled per query.
The maximum number of turns is set to 10, the penalty scale $\lambda$ is set to 0, and the discount factor $\gamma$ is set to 0.9.
The reinforcement learning phase spans 3 epochs, conducted on 8 NVIDIA A800-80G GPUs.
Please refer to Appendix~\ref{sec:appendix_implementation} for more details about hyperparameter configurations.

\subsection{Experimental Results}
We present the experimental results in Table~\ref{tab:main}, from which we derive the following observations:

$\bullet$~Firstly, compared to vanilla models, most post-training methods generally achieve consistent performance gains across all benchmarks, demonstrating the effectiveness of reinforcement learning for optimizing complex agentic behaviors in TIR.

$\bullet$~Secondly, compared to single-turn training, models trained in a multi-turn setting exhibit significantly superior performance under similar reward designs, confirming the necessity of end-to-end optimization for complex interaction sequences.

$\bullet$~Furthermore, compared to all baseline methods, both variants of \ourmodel achieve the highest performance across three benchmarks, demonstrating strong generalization that overcomes the potential exploration constraints imposed by ground-truth alignment.
Notably, our 4B models even surpass most 8B baselines.
This improvement can be attributed to our turn-level reward design and dual-level advantage estimation, which effectively distinguishes useful tool calls from mistakes.

$\bullet$~Finally, compared to the soft variant (OT), the hard variant (KM) consistently yields better results.
This suggests that soft alignment may assign partial credit to near-miss tool calls (e.g., parameter errors) which are actually fatal in execution, leading to suboptimal policy updates.
In contrast, KM provides a rigorous binary signal essential for exact tool usage by enforcing a strict one-to-one mapping.

\subsection{Ablation Study}

\begin{table}[t]
	% \small
	\centering
        \resizebox{1\linewidth}{!}{
	\begin{tabular}{lcccc}
		\toprule
		\multirow{2.5}{*}{\textbf{Reward Design}} & \multicolumn{4}{c}{\textbf{FTRL}} 
        \\ 
		\cmidrule(lr){2-5}
		& Solve-P & Solve-R & Solve-F1 & Avg.\\ 
		\hline
		\rowcolor{Gainsboro} \multicolumn{5}{c}
        {\textit{\textbf{Trajectory-Level Advantage Estimation}}} \\
        \midrule    
            \textbf{Outcome} &31.59 & 39.75 & 32.54 & 34.63 \\
            \textbf{Turn-Level} &32.70 & 40.70 & 35.08 & 36.16 \\
            \textbf{Integrated Reward} &32.31 & 43.16 & 35.11 & 36.86 \\
        \hline
        \rowcolor{Gainsboro} \multicolumn{5}{c}{\textit{\textbf{Turn-Level Advantage Estimation}}}\\
        \midrule
        \textbf{Turn-Level} &29.50 & 42.06 & 32.01 & 34.52 \\
        \textbf{Integrated Reward} &32.88 & 41.67 & 34.65 & 36.40 \\
        \hline
        \rowcolor{Gainsboro} \multicolumn{5}{c}{\textit{\textbf{Dual-Level Advantage Estimation}}}\\
        \midrule
        \textbf{Turn-Level} &32.45 &43.08  &35.06 &36.96 \\
        \cellcolor{ablationblue}\textbf{\ourmodel} &\cellcolor{ablationblue}\textbf{36.33} & \cellcolor{ablationblue}\textbf{44.18} & \cellcolor{ablationblue}\textbf{37.33} & \cellcolor{ablationblue}\textbf{39.28} \\
		\bottomrule
	\end{tabular}
	}
    \vspace{-0.2cm}
	\caption{Ablation study on reward design and advantage estimation on the FTRL dataset using Qwen3-8B.}
        \vspace{-0.6cm}
	\label{tab:ab}
\end{table}

To evaluate the contribution of each component within our framework, we conduct a comprehensive ablation study on the Qwen3-8B model across two dimensions: \textbf{reward design} and \textbf{advantage estimation strategies}.
As shown in Table~\ref{tab:ab}, turn-level rewards consistently outperform outcome-based rewards, indicating the importance of supervising intermediate tool interaction steps.
Furthermore, combining turn-level and outcome rewards yields additional performance gains, indicating that effective optimization for TIR benefits from jointly modeling intermediate tool execution quality and final task correctness.
Regarding advantage estimation, using either trajectory-level or turn-level advantage alone leads to suboptimal performance.
In contrast, integrating both consistently yields superior results across all evaluation metrics. 
This suggests that assigning token-level credit requires balancing fine-grained local execution signals with global trajectory-level context.
Overall, \ourmodel achieves the best performance by effectively synthesizing these components, 
confirming that fine-grained turn-level signals effectively complement global trajectory-level supervision, enabling more precise and reliable optimization.

\subsection{Robustness to Task Complexity}
\begin{figure}[t]
    \centering
    \subfigtopskip=2pt 
	\subfigbottomskip=2pt 
	\subfigcapskip=-5pt 
    \subfigure{
    \includegraphics[width=0.465\linewidth]{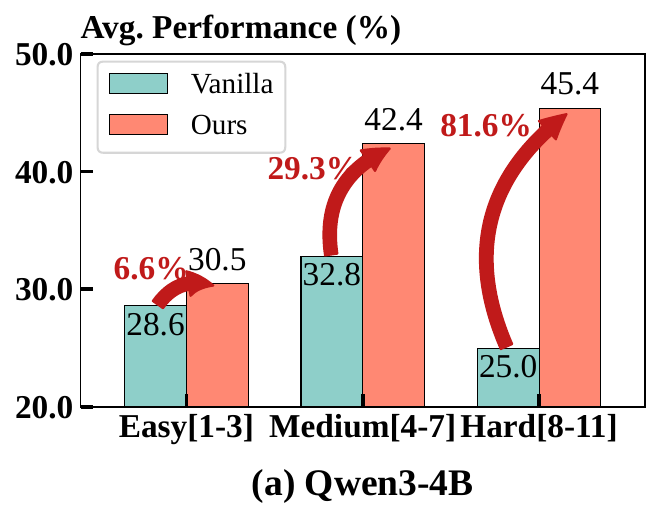}
    }
    \subfigure{
    \includegraphics[width=0.465\linewidth]{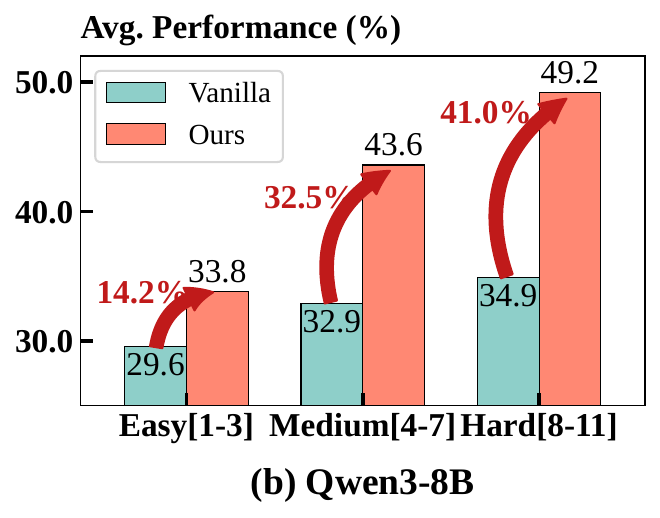}
    }
    \vspace{-0.8cm}
    \caption{Performance comparison on subsets of varying task complexity, where queries are grouped based on the number of tools required to solve each task.}
    % \vspace{-1em}
    \label{fig:analy_group}
    \vspace{-0.2cm}
\end{figure}

To evaluate how \ourmodel adapts to varying reasoning depths, we categorize the evaluation queries into three difficulty levels, namely Easy, Medium, and Hard, based on the number of tool invocations required to resolve each task.
As illustrated in Figure~\ref{fig:analy_group}, \ourmodel consistently outperforms the vanilla baseline across all complexity levels for both the 4B and 8B model scales, demonstrating robust generalizability.
Notably, the performance gains become increasingly pronounced as task complexity scales, with the most significant gains observed on the Hard subset.
This trend indicates that long-horizon, multi-turn tool-interaction tasks benefit more substantially from turn-level reward assignment.
In these complex scenarios, individual tool calls contribute unevenly to the final outcome, and a uniform advantage to all tokens within such a long trajectory inevitably obscures these critical distinctions.
By providing fine-grained supervision, \ourmodel enables more effective credit assignment, thereby yielding superior performance, especially on long-horizon, multi-turn tasks.

\subsection{Tool-Use Efficiency and Accuracy}

\begin{table}[t]
\centering
\resizebox{0.95\linewidth}{!}{
% \small
\begin{tabular}{lcccc}
\toprule
\textbf{Model} & \textbf{Method} & \textbf{Tool Calls $\downarrow$} & \textbf{Success $\uparrow$} & \textbf{Failure $\downarrow$} \\ 
\midrule
\multirow{2}{*}{\textbf{Qwen3-4B}} & Vanilla & 1444 & 15.44\% & 84.56\% \\
                           & \textbf{Ours} & \textbf{1297} & \textbf{27.83\%} & \textbf{72.17\%} \\
\midrule
\multirow{2}{*}{\textbf{Qwen3-8B}} & Vanilla & 1593 & 19.08\% & 80.92\% \\
                           & \textbf{Ours} & \textbf{1373} & \textbf{27.90\%} & \textbf{72.10\%} \\
\bottomrule
\end{tabular}
}
\vspace{-0.2cm}
\caption{Analysis of tool-use efficiency and accuracy.}
\label{tab:agentic_behavior}
\vspace{-0.4cm}
\end{table}
To assess how turn-level reward assignment influences the tool-use behavior of the agent, we analyze the total number of tool calls as well as the proportions of successful and failed tool invocations.
As shown in Table~\ref{tab:agentic_behavior}, models trained with our approach consistently make fewer tool calls while achieving substantially higher tool-call success rates across both Qwen3-4B and Qwen3-8B.
Specifically, our method reduces the overall frequency of tool invocations and significantly lowers the failure rate, indicating that agents learn to invoke tools more selectively and accurately rather than relying on excessive or redundant calls.
These results suggest that fine-grained, turn-level supervision not only improves task-level performance but also leads to more efficient and reliable tool-use strategies.

\subsection{Hyper-parameter Analysis}
\begin{figure}[t]
    \centering
    \subfigtopskip=2pt 
	\subfigbottomskip=2pt 
	\subfigcapskip=-5pt 
    \subfigure{
    \includegraphics[width=0.465\linewidth]{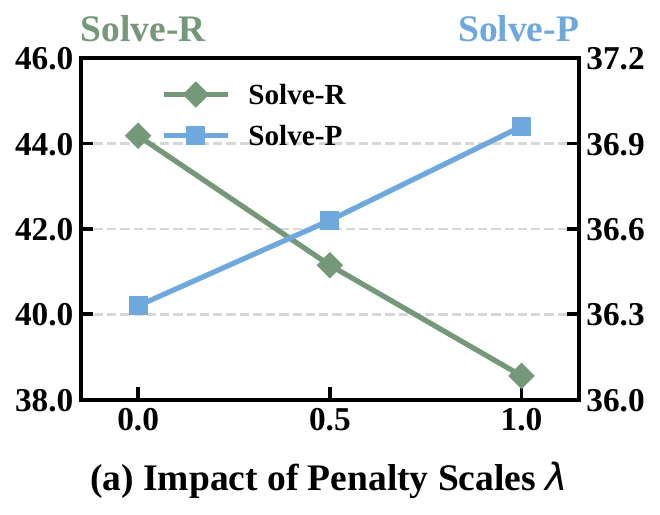}
    }
    \subfigure{
    \includegraphics[width=0.465\linewidth]{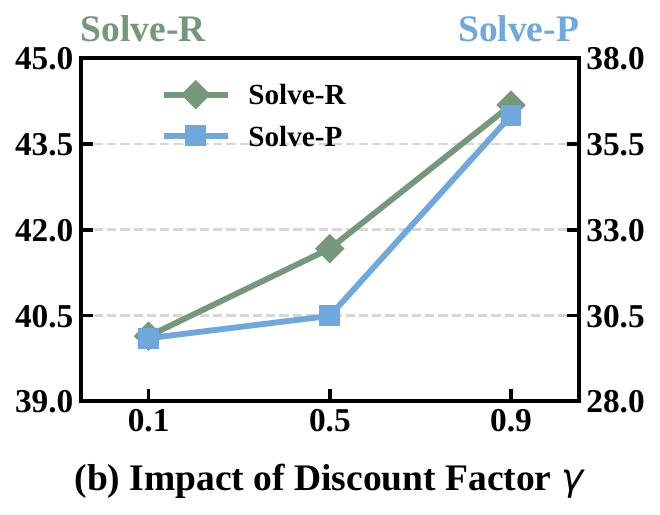}
    }
    \vspace{-0.6cm}
    \caption{Sensitivity analysis of performance to hyper-parameters. (a) shows the impact of penalty scales $\lambda$. (b) illustrates the effect of discounted factor $\gamma$.}
    \vspace{-0.6cm}
    \label{fig:analy_hyper}
\end{figure}

\paratitle{Impact of Penalty Scales.}
As illustrated in Figure~\ref{fig:analy_hyper}(a), we observe a distinct trade-off between precision and success rate as the penalty scale $\lambda$ increases.
Specifically, intensifying the penalty for unmatched or redundant tool calls leads to a consistent improvement in Solve-P, indicating that the model becomes more conservative and effectively suppresses redundant or erroneous tool invocations.
However, this increased stringency results in a marginal decline in Solve-R, suggesting that overly strong penalties may inhibit exploratory yet necessary tool calls in high-uncertainty scenarios.
These results highlight the importance of balancing penalty strength to encourage precise tool usage without excessively constraining the model’s ability to explore and resolve complex queries.

\paratitle{Impact of Discount Factor.}
Figure~\ref{fig:analy_hyper}(b) demonstrates the performance sensitivity to the discount factor $\gamma$. 
We observe a consistent and significant improvement across all evaluation metrics as $\gamma$ increases from $0.1$ to $0.9$.
This trend underscores the long-horizon nature of multi-turn TIR, where early tool-calling decisions critically shape subsequent interactions and ultimately determine task success.
A higher $\gamma$ effectively enables the turn-level advantage to capture these long-term dependencies, ensuring that intermediate actions are optimized not just for immediate correctness, but for their contribution to the overall task completion.

\section{Related Work}
% 1.TIR
% 2.Credit Assignment
\subsection{Tool-Integrated Reasoning}
% 这一段讲TIR然后说他们都是trajectory level的reward
% Tool Integrated Reasoning (TIR) has emerged as a promising paradigm for enhancing the capabilities of LLMs by enabling them to interact with external tools 早期工作通过从lrm中蒸馏出轨迹然后通过sft使得模型可以边推理边使用工具，但是这种方法泛化性较差，近期通过GRPO等RLVR方法增强模型TIR能力吸引了大量关注，相比于sft进一步提高了模型的能力与泛化性。despite these advances，大多数方法仅assign 整个trajectory一个reward，where all actions within a trajectory share the same reward regardless of quality ，Such coarse feedback prevents the model from distinguishing between effective and ineffective tool-calling。in this paper，we propose a method to assign turn-level reward，
% 缩短空间的话可以不讲SFT 直接讲RL的
% Tool-Integrated Reasoning (TIR) has emerged as a promising approach to enhance the capabilities of LLMs by enabling them to interact with external tools during the reasoning process. 
% Early approaches primarily rely on distilling TIR trajectories from LRMs and training models via supervised fine-tuning (SFT), which often exhibit limited generalization beyond seen tools and reasoning patterns.
% More recently, reinforcement learning with verifiable rewards (RLVR), such as GRPO-based methods, has shown improved reasoning performance and generalization by optimizing models against task-level feedback.
Recently, RLVR, such as GRPO-based methods \cite{shao2024deepseekmath,yu2025dapo,zheng2025group}, has been widely adopted to enhance TIR by directly optimizing models against task-level feedback~\cite{wang2025otc,xue2025simpletir,qu2025tool}. 
Early RL-based approaches rely on rewards derived solely from final outcomes, resulting in sparse and delayed supervision~\cite{li2025torl,feng2025retool}. 
Subsequent methods mitigate this issue by assigning rewards at the trajectory level, providing denser training signals~\cite{qian2025toolrl,zhang2025nemotron,zeng2025tool,wei2025autotir}.
Despite these advances, most existing methods assign a single reward to an entire trajectory, causing all actions to share identical advantage signals regardless of their individual quality. 
Such coarse-grained feedback prevents models from distinguishing effective from ineffective tool calls and obscures turn-level contributions in multi-turn scenarios~\cite{lightman2024lets}.
In contrast, our work introduces a turn-level reward assignment and dual-level advantage estimation framework, which enables fine-grained supervision.

\subsection{Fine-Grained Credit Assignment}
% 这一段说
Recent studies have explored turn-level or process-level rewards to perform fine-grained credit assignment~\cite{zeng2025reinforcing,yu2025steptool,yue2025promoting,zheng-etal-2025-stepsearch,zhang2025process}, which can be broadly categorized into sampling-based and model-based methods.
Sampling-based approaches typically employ Monte Carlo rollouts to approximate turn-level credit~\cite{feng2025groupingroup,tran2025exploiting}.
While theoretically unbiased, these approaches suffer from high variance and substantial computational overhead, especially in multi-turn interaction settings.
Model-based methods instead leverage intrinsic reward functions or external reward models to estimate the utility of intermediate steps~\cite{wang2025information,wang2025spa}. 
However, the accuracy of these rewards is not guaranteed, as these signals are often model-dependent, difficult to calibrate, and prone to limited generalization across domains.
In this paper, we exploit the structured  nature of TIR by formulating turn-level credit assignment as a bipartite matching problem, enabling fine-grained and precise turn-level rewards.

\section{Conclusion}
In this paper, we propose \ourmodel, a novel framework designed to provide fine-grained supervision for Tool-integrated Reasoning (TIR) via bipartite matching.
By introducing a turn-level reward design and dual-level advantage estimation, our method assigns distinct advantages to individual tokens, effectively distinguishing high-quality tool calls from redundant or uninformative ones.
Extensive experiments on both in-domain and out-of-domain benchmarks demonstrate the effectiveness of \ourmodel across various model scales, with particularly pronounced improvements in long-horizon, multi-turn interaction scenarios.

\section*{Limitations}
Despite the significant performance gains achieved on the Qwen3-4B and Qwen3-8B model series, our work has certain limitations.
First, due to constrained computational resources, we were unable to conduct experiments on larger-scale models.
We hope that the publication of this work will encourage further exploration of our framework on more massive architectures.
Second, our approach relies on the availability of ground-truth trajectories for turn-level supervision. 
This poses a challenge in highly open-ended scenarios such as ``deep research'' tasks, where ground-truth parameters and reasoning steps are inherently difficult to determine or verify. 
Consequently, adapting our method to such problems remains a promising but challenging direction for future research.

\bibliography{custom}

\clearpage
\appendix

\begin{table*}[h]
\centering
\resizebox{1.0\textwidth}{!}{
\begin{tabular}{l|c|cccc|cccccc|c}
\toprule
\multirow{2}{*}{\textbf{Datasets}} & \textbf{Train} & \multicolumn{11}{c}{\textbf{Test}} \\ \cmidrule{2-13}
 & \multirow{2}{*}{FTRL} & \multicolumn{4}{c|}{FTRL} & \multicolumn{6}{c|}{BFCL} & \multirow{2}{*}{ToolHop} \\ \cmidrule{3-12}
 &  & Single-Hop & Parallel SH & Multi-Hop & Parallel MH & Base & Miss Func & Miss Param & Long Context & Web Search & Memory &  \\ \midrule
\#Number & 2,215 & 50 & 50 & 50 & 50 & 200 & 200 & 200 & 200 & 200 & 465 & 995 \\ \bottomrule
\end{tabular}
}
\caption{Detailed statistics of the experimental datasets, including training data from FTRL and test sets across FTRL, BFCL, and ToolHop benchmarks.}
\label{appendix:tab_statistic}
\end{table*}

\section*{Appendix}

\section{Benchmark Details}
\label{sec:appendix_benchmark}
In this section, we introduce the detailed description of the benchmarks used in this paper. 
The statistics of these datasets are shown in Table~\ref{appendix:tab_statistic}.

\paratitle{FTRL.}\footnote{{\href{https://github.com/bytedance/FTRL}{https://github.com/bytedance/FTRL}}} The Feedback-Driven Tool-Use Reinforcement Learning (FTRL)~\cite{ye2025feedback} benchmark is designed for evaluating tool-integrated reasoning under automatically constructed local execution environments.
To mirror the complexity of real-world interactions, FTRL categorizes tool-use scenarios into four distinct levels of difficulty: 

$\bullet$~\textbf{Single-Hop}: Contains a single query that can be resolved with a one-time tool invocation. 

$\bullet$~\textbf{Parallel Single-Hop}: Involves multiple independent sub-queries, allowing the agent to invoke multiple tools in parallel to gather information. 

$\bullet$~\textbf{Multi-Hop}: Consists of a sequence of dependent sub-queries, where the execution of a subsequent tool depends on the output of previous steps. 

$\bullet$~\textbf{Parallel Multi-Hop}: A hybrid and complex structure that includes both independent and interdependent sub-queries, requiring advanced logical planning and orchestration.

To provide a multidimensional assessment of model performance, FTRL employs the following verifiable metrics:

$\bullet$~\textbf{Solve-P}: Measures the precision of the tool invocations. It is defined as the ratio of valid tool calls to the total number of calls made:
\begin{equation}
\text{Solve-P} = \begin{cases}
\frac{q}{p}, & \text{if}\quad p>0 \\
1, & \text{if} \quad p=0
\end{cases}
\end{equation}
where $p$ is the number of tools invoked, $q$ is the number of sub-questions successfully solved.

$\bullet$~\textbf{Solve-R}: Evaluates the completeness of task execution by measuring the proportion of successfully resolved sub-tasks. The formula is:
\begin{equation}
    \text{Solve-R} = \frac{q}{n}
\end{equation}
where n is the number of sub-questions in total.

\textbf{Solve-F1}: Evaluates the harmonic mean of Solve-P and Solve-R, providing an overall performance measure:
\begin{equation}
    \text{Solve-F1} = \frac{2\cdot \text{Solve-P} \cdot \text{Solve-R}}{\text{Solve-P} +\text{Solve-R}}
\end{equation}

In this work, we use FTRL as the in-domain benchmark for both training and evaluation.
Its structured feedback and multi-turn interaction setting make it particularly suitable for studying fine-grained credit assignment and turn-level supervision in tool-integrated reasoning.

\paratitle{BFCL.}\footnote{{\href{https://github.com/ShishirPatil/gorilla/blob/main/berkeley-function-call-leaderboard/README.md}{https://github.com/ShishirPatil/gorilla/blob/main/berkeley-function-call-leaderboard}}}
The Berkeley Function Calling Leaderboard (BFCL)~\cite{patil2025bfcl} is a comprehensive and rigorous benchmark designed to evaluate the function-calling capabilities of LLMs across a wide range of scenarios.
To assess the performance of models in complex agentic settings, BFCL V3 and V4 introduce several critical task dimensions:

$\bullet$~\textbf{Multi-Turn}:
This dimension evaluates the model's ability to maintain state and handle dynamic decision-making over multiple conversational turns. It consists of four subsets: \textbf{Base}, which provides standard multi-turn interactions; \textbf{Missing Parameters} and \textbf{Missing Functions}, which test robustness when information is incomplete; and \textbf{Long-Context}, which requires extracting tool information from extensive distractor text.

$\bullet$~\textbf{Agentic}:
This dimension focuses on the model's capacity for autonomous reasoning and interaction within a stateful environment. it includes two subsets: \textbf{Web Search}, requiring real-time information retrieval, and \textbf{Memory}, testing the ability to recall and utilize information from previous interactions.

BFCL employs a multi-layered evaluation strategy to ensure execution rigor: 

$\bullet$~\textbf{Multi-turn}: Requires passing two per-turn checks for overall success: (1) \textbf{State-based Evaluation}, which verifies backend system state changes, and (2) \textbf{Response-based Evaluation}, which validates the execution path against minimal viable result paths for read-only requests. 

$\bullet$~\textbf{Agentic}: Employs an exact-match metric isolated to the ``answer'' field. This prevents false positives from incidental phrasing in unstructured text, ensuring the model commits to a precise and intended response.

\paratitle{ToolHop.}\footnote{{\href{https://huggingface.co/datasets/bytedance-research/ToolHop}{https://huggingface.co/datasets/bytedance-research/ToolHop}}}
ToolHop~\cite{ye2025toolhop} is a query-driven benchmark specifically designed to evaluate LLMs in multi-hop tool-use scenarios. It features 995 user queries and 3,912 locally executable tools, focusing on the model's ability to handle complex interdependencies.

ToolHop focuses on the accuracy of the final answer through a verifiable execution environment:
\textbf{Answer Correctness (AC)}: The primary metric that measures the proportion of queries for which the model provides a correct final answer, verified against ground-truth labels.

\section{More Implementation Details}
\label{sec:appendix_implementation}
The detailed hyper-parameter configurations for our experiments are summarized in Table~\ref{tab:training_hparams}. 
\begin{table}[ht]
\centering
\small
\begin{tabular}{lc}
\toprule
\textbf{Category} & \textbf{Hyper-parameter} \\
\midrule
\multicolumn{2}{l}{\textbf{Data Configuration}} \\
\midrule
Train Batch Size & 256 \\
Validation Batch Size & 256 \\
Max Prompt Length & 7000 \\
Max Response Length & 23000 \\
\midrule
\multicolumn{2}{l}{\textbf{Optimization}} \\
\midrule
Learning Rate & 1e-6 \\
PPO Mini Batch Size & 32 \\
KL-Divergence loss coefficient & 0.001 \\
\midrule
\multicolumn{2}{l}{\textbf{Rollout Configuration}} \\
\midrule
Rollout Name & vllm \\
Rollout Temperature & 1.0 \\
GPU Memory Utilization & 0.7 \\
Number of Rollouts & 16 \\
\midrule
\multicolumn{2}{l}{\textbf{Training \& Logging}} \\
\midrule
Save Frequency & 8 \\
Test Frequency & 4 \\
Total Epochs & 3 \\
\bottomrule
\end{tabular}
\caption{Training Configuration.}
\label{tab:training_hparams}
\end{table}

In addition to the standard training setup, we provide further clarifications on the implementation of our reward and evaluation mechanisms.
Our turn-level advantage estimation is performed by comparing rollouts within the same group at each specific turn $t$. Given that trajectories in a group may have varying lengths, we implement the following logic to ensure robust estimation:
For a given turn $t$, the advantage is computed only among the subset of rollouts that actually reach or exceed $t$ turns.
In scenarios where only a single rollout within a group contains the $t$-th turn, we provide a fallback by setting the mean to 0 and the variance to 1 for normalization purposes. 
This prevents numerical instability and ensures that the single available data point does not lead to skewed gradients.
For the evaluation of the \textbf{Web Search} subset within the BFCL benchmark, we utilize the Bing Search~\footnote{{\href{https://serpapi.com/bing-search-api}{https://serpapi.com/bing-search-api}}} API as the underlying search engine.

\section{More Experiments}
\label{sec:appendix_experiments}
We further perform additional experiments on FTRL with Qwen3-8B for deeper analysis, which are presented here due to space constraints.

\begin{table}[t]
	\centering
        \resizebox{1\linewidth}{!}{
	\begin{tabular}{lcccc}
		\toprule
		\multirow{2.5}{*}{\textbf{Advantage Estimation}} & \multicolumn{4}{c}{\textbf{FTRL}} 
        \\ 
		\cmidrule(lr){2-5}
		& Solve-P & Solve-R & Solve-F1 & Avg.\\ 
        \midrule    
            \textbf{Weighted Product (KM)} &33.10 & 42.38 & 34.62 & 36.70 \\
            \textbf{Weighted Sum (KM)} &29.95 & 40.71 & 32.77 & 34.48 \\
            \textbf{\ourmodel (KM)} &\textbf{36.33} & \textbf{44.18} & \textbf{37.33} & \textbf{39.28} \\
		\bottomrule
	\end{tabular}
	}
    % \vspace{-0.2cm}
	\caption{Additional analysis on advantage estimation on the FTRL dataset using Qwen3-8B.}
        % \vspace{-0.6cm}
	\label{appendix:tab_advantage}
\end{table}

\subsection{Analysis on Advantage Estimation}
To further investigate the mechanism of turn-level credit assignment, we conduct supplementary experiments using the Qwen3-8B model on the FTRL benchmark.
We compare \ourmodel with an alternative intuition-based approach, referred to here as the Weighted Intra-trajectory Method.

\paratitle{Weighted Intra-trajectory Method.}
To further explore the mechanisms of turn-level credit assignment, we investigate two alternative variants that estimate advantages within an individual trajectory (intra-trajectory) rather than across different rollouts (inter-group). 
For both variants, we first compute a local turn-level advantage $A_{i,t}^{loc}$ by treating each turn as an independent sample within the same rollout:
$A_{i,t}^{loc} = \frac{R_{i,t} - \text{mean}(\{R_{i,j}\}_{j=1}^{T})}{\text{std}(\{R{i,j}\}_{j=1}^{T})}.$ 
Based on $A_{i,t}^{loc}$, we propose two weighting strategies to adjust the global trajectory advantage $A_i^g$:

$\bullet$~\textbf{Weighted Product}: This method uses the local advantage as a scaling factor to modulate the magnitude of the global signal.
The final adjusted advantage $\hat{A}_{i,t}$ for each token in turn $t$ is defined as
$\hat{A}_{i,t} = \left( 1 + 0.1 \cdot\text{sgn}(A_i^g) \cdot A_{i,t}^{loc} \right) A_i^g$, 
where $\text{sgn}(\cdot)$ ensures the weighting direction aligns with the trajectory performance.

$\bullet$~\textbf{Weighted Sum}: This method directly incorporates the local turn-level signal by adding it to the global advantage:
$\hat{A}_{i,t} = A_i^g + A_{i,t}^{loc}$

\paratitle{Results and Discussion.} 
As shown in Table~\ref{appendix:tab_advantage}, the results indicate that this Weighted Intra-trajectory Method approach yields inferior performance compared to \ourmodel.
We attribute this to the fact that intra-trajectory normalization inherently overlooks the temporal interdependencies between turns. 
In multi-turn TIR tasks, the success of a later turn is often contingent upon the precise execution of earlier tool calls. 
By solely focusing on the internal variance of a single rollout, the weighted method fails to capture the relative quality of the reasoning path across different potential strategies, whereas our inter-group discounted approach provides a higher-resolution signal for mastering long-horizon interactions.

\begin{table}[t]
	% \small
	\centering
        \resizebox{1\linewidth}{!}{
	\begin{tabular}{lcccc}
		\toprule
		\multirow{2.5}{*}{\textbf{Cost Matrix}} & \multicolumn{4}{c}{\textbf{FTRL}} 
        \\ 
		\cmidrule(lr){2-5}
		& Solve-P & Solve-R & Solve-F1 & Avg.\\ 
        \midrule    
            \textbf{Exponential Transformation} &30.96 & 41.50 & 33.38 & 35.28 \\
             \textbf{Normalized Inversion} &32.77 & \textbf{42.88} & \textbf{34.91} & \textbf{36.85} \\
            \textbf{Linear Inversion} &\textbf{33.61} & 42.56 & 33.61 & 36.59 \\
		\bottomrule
	\end{tabular}
	}
	\caption{Additional analysis on cost matrix construction of OT on the FTRL dataset using Qwen3-8B.}
	\label{appendix:tab_cost}
\end{table}

\subsection{Analysis on Cost Matrix Construction}
Since the cost matrix $C$ in our Optimal Transport (OT) framework is derived from the similarity matrix $S$, the transformation function applied to $S$ significantly influences the final turn-level reward assignment. 
To evaluate this impact, we compare three distinct construction methods:

$\bullet$~\textbf{Linear Inversion}: $C = -S$. This maintains the original linear distribution of similarity scores.

$\bullet$~\textbf{Normalized Inversion}: $C = 1 - S_{\text{norm}}$, where $S_{\text{norm}} = \frac{S - \min(S)}{\max(S) - \min(S) + \epsilon}$. 

$\bullet$~\textbf{Exponential Transformation}: $C = -e^S$. This non-linear mapping amplifies the differences between high-similarity pairs and others.

\paratitle{Results and Discussion.} 
As shown in Table~\ref{appendix:tab_cost}, the results indicate that the model performance is relatively insensitive to the choice between Linear Inversion and Normalized Inversion. This consistency suggests that the raw similarity scores generated by our matching module are already well-distributed, and linear re-scaling does not alter the relative matching order.
In contrast, we observe a performance drop with the Exponential Transformation. 
We hypothesize that this is primarily due to numerical instability and noise amplification. 
The exponential function excessively magnifies minor fluctuations in similarity scores, creating a cost matrix with values spanning several orders of magnitude. 
This not only hinders the convergence of the Sinkhorn solver due to precision errors during iterative normalization but also leads to high-variance advantage that destabilize the policy gradient. 

\begin{table}[t]
	% \small
	\centering
        \resizebox{1\linewidth}{!}{
	\begin{tabular}{lcccc}
		\toprule
		\multirow{2.5}{*}{\textbf{Methods}} & \multicolumn{4}{c}{\textbf{FTRL}} 
        \\ 
		\cmidrule(lr){2-5}
		& Solve-P & Solve-R & Solve-F1 & Avg.\\ 
        \hline
        \rowcolor{Gainsboro} \multicolumn{5}{c}{\textit{\textbf{FTRL}}}\\
        \midrule  
            \textbf{Single-Turn} &32.32 & 38.87 & 32.85 & 34.68 \\
            \textbf{Expanded Single-Turn} &\textbf{35.34} & \textbf{43.13} & \textbf{35.30} & \textbf{37.92} \\
            \textbf{Multi-Turn} &33.28 & 40.60 & 34.20 & 36.03 \\
        \hline
        \rowcolor{Gainsboro} \multicolumn{5}{c}{\textit{\textbf{ToolRL}}}\\
        \midrule  
            \textbf{Single-Turn} &25.57 & 35.31 & 26.72 & 29.20 \\
            \textbf{Expanded Single-Turn} &30.23 & 27.73 & 27.28 & 28.41 \\
            \textbf{Multi-Turn} &\textbf{30.64} & \textbf{40.07} & \textbf{32.51} & \textbf{34.41} \\
		\hline
        \rowcolor{Gainsboro} \multicolumn{5}{c}{\textit{\textbf{\ourmodel}}}\\
        \midrule 
        \cellcolor{ablationblue}\textbf{Multi-Turn} &\cellcolor{ablationblue}\textbf{36.33} & \cellcolor{ablationblue}\textbf{44.18} & \cellcolor{ablationblue}\textbf{37.33} & \cellcolor{ablationblue}\textbf{39.28} \\
		\bottomrule
	\end{tabular}
	}
    % \vspace{-0.2cm}
	\caption{Performance comparison between multi-turn training and expanded single-turn baselines on FTRL and ToolRL using Qwen3-8B.}
        % \vspace{-0.6cm}
	\label{appendix:tab_single_turn}
\end{table}

\subsection{Multi-Turn vs. Expanded Single-Turn}
While our main results demonstrate the superiority of multi-turn training, one could argue that this advantage stems from increased computational expenditure during inference and training. 
To ensure a fair comparison, we conduct an experiment by decomposing multi-turn trajectories into expanded single-turn training sets.
Specifically, a trajectory consisting of $T$ turns is decomposed into $T-1$ independent training samples. 
For each sample at turn $t$, the model is provided with the conversation history of the preceding $t$ turns and tasked with predicting the $(t+1)$-th tool-calling action. 
This data expansion ensures that the single-turn baseline is exposed to the same volume of state-transition information as the multi-turn approach.

\paratitle{Results and Discussion.} As shown in Table~\ref{appendix:tab_single_turn}, while increasing the data volume via decomposition improves performance on the \textbf{FTRL} benchmark, it still falls short of our end-to-end multi-turn optimization. 
Interestingly, on the \textbf{ToolRL} benchmark, we observe that increasing data through single-turn decomposition actually leads to a performance decline.
We attribute this to the inherent limitations of the single-turn paradigm: Single-turn models rely heavily on the immediate reward design for each isolated step. 
Without the context of the entire trajectory, they struggle to reconcile local actions with long-term goals.
The single-turn format restricts the potential for sophisticated reward and advantage estimation, which requires a holistic view of the interaction.
These findings suggest that end-to-end multi-turn optimization is fundamentally more effective, as it allows the model to capture inter-turn dependencies that simple data expansion in a single-turn setting cannot replicate.
\section{Case Study}
\label{sec:appendix_case}
In this section, we present a case study to demonstrate the rationality of the turn-level rewards generated by \ourmodel. 
As shown in Table~\ref{appendix:tab_case_study}, this example specifically highlights the distinction between the hard and soft variants.

\begin{table*}[!htb]
\centering \footnotesize
\begin{tabular}{p{49em}}
\toprule
\textbf{{\textbf{Question}:}} What is the common building material of the town where the architect who designs buildings inspired by the architecture of the castle on the hill overlooking the valley? \\ {\textbf{Ground Truth Tool Calls:}} 
[\{ ``name'': ``valley\_hill\_analyzer'', ``parameters'': \{\} \},  \{ ``name'': ``landmark\_locator'', ``parameters'': \{``landmark\_type'': ``castle'', ``geographic\_feature'': ``Yosemite Valley'', ``position\_relation'': ``on the hill''\} \}, \{ ``name'': ``architect\_inspiration\_analyzer'', ``parameters'': \{``landmark'': ``hearst castle''\} \}, \{ ``name'': ``person\_locator'', ``parameters'': \{``name'': ``Lily Chen''\} \}, \{ ``name'': ``building\_material\_analyzer'', ``parameters'': \{``location'': ``Santa Barbara''\} \} ]
\\ {\textbf{Ground Truth Answer:}} Stone.
\\ \hline
\vspace{-4pt}
{\textbf{Turn 1}:}\\ <think> Okay, let me try to figure out how to answer this question. ...... </think>\\
<tool\_call> \{``name'': ``landmark\_locator'', ``arguments'': \{``landmark\_type'': ``castle'', ``geographic\_feature'': ``\textcolor{red}{valley}'', ``position\_relation'': ``\textcolor{red}{overlooking}''\}\} </tool\_call>\\
<tool\_response> \{``result'': ``No specific landmark found matching the criteria''\} </tool\_response>\\
\textcolor{red}{[WRONG]}\space \space\space\space \space\space \space \space\space\space \space\space \space \space\space\space \space\space \space \space\space\space  \textcolor{purple}{\textbf{Reward Given by \ourmodel(KM): 0}}\space \space\space\space \space\space \space \space\space\space \space\space \space \space\space\space \space\space \space \space \space\space\textcolor{blue}{\textbf{Reward Given by \ourmodel(OT):}} \textcolor{red}{\textbf{0.025}}
\\ \hline
\vspace{-3pt}
{\textbf{Turn 2}:}\\ <think> Okay, the first tool call didn't return a specific landmark. Maybe I need to ...... </think>\\
<tool\_call> \{``name'': ``valley\_hill\_analyzer'', ``arguments'': \{\}\} </tool\_call>\\
<tool\_response> \{``result'': ``Yosemite Valley has a hill overlooking it.''\}</tool\_response>\\
\textcolor{DarkGreen}{[CORRECT]} \space\space\space\space\space\space \space\space \space \space\space\space \space\space \space \space\space\space \space\space \textcolor{purple}{\textbf{Reward Given by \ourmodel(KM): 1.0}}\space\space \space\space \space \space\space\space \space\space \space \space\space\space \space\space \space \space\space\space \space\space \textcolor{blue}{\textbf{Reward Given by \ourmodel(OT): 0.1667}}
\\ \hline
\vspace{-3pt}
{\textbf{Turn 3}:}\\ <think> Okay, so the valley\_hill\_analyzer returned Yosemite Valley with a hill overlooking it. Now, the next step is to ...... </think>\\
<tool\_call> \{``name'': ``landmark\_locator'', ``arguments'': \{``landmark\_type'': ``castle'', ``geographic\_feature'': ``Yosemite Valley'', ``position\_relation'': ``on the hill''\}\} </tool\_call>\\
<tool\_response> \{``result'': ``Hearst Castle''\}</tool\_response>\\
\textcolor{DarkGreen}{[CORRECT]} \space\space\space\space\space\space \space\space \space \space\space\space \space\space \space \space\space\space \space\space \textcolor{purple}{\textbf{Reward Given by \ourmodel(KM): 1.0}}\space\space \space\space \space \space\space\space \space\space \space \space\space\space \space\space \space \space\space\space \space\space \textcolor{blue}{\textbf{Reward Given by \ourmodel(OT):}} \textcolor{red}{\textbf{0.1581}}
\\ \hline
\vspace{-3pt}
{\textbf{Turn 4}:}\\ <think> Okay, so the Hearst Castle is located on the hill overlooking Yosemite Valley. Now, the next step is to ...... </think>\\
<tool\_call> {``name'': ``architect\_inspiration\_analyzer'', ``arguments'': {``landmark'': ``Hearst Castle''}} </tool\_call>\\
<tool\_response> Lily Chen </tool\_response>\\
\textcolor{DarkGreen}{[CORRECT]} \space\space\space\space\space\space \space\space \space \space\space\space \space\space \space \space\space\space \space\space \textcolor{purple}{\textbf{Reward Given by \ourmodel(KM): 1.0}}\space\space \space\space \space \space\space\space \space\space \space \space\space\space \space\space \space \space\space\space \space\space \textcolor{blue}{\textbf{Reward Given by \ourmodel(OT): 0.1667}}
\\ \hline
\vspace{-3pt}
{\textbf{Turn 5}:}\\ <think> Okay, so the architect inspired by Hearst Castle is Lily Chen. Now, I need to find out ...... </think>\\
<tool\_call> {``name'': ``person\_locator'', ``arguments'': {``name'': ``Lily Chen''}} </tool\_call>\\
<tool\_response> Santa Barbara </tool\_response>\\
\textcolor{DarkGreen}{[CORRECT]} \space\space\space\space\space\space \space\space \space \space\space\space \space\space \space \space\space\space \space\space \textcolor{purple}{\textbf{Reward Given by \ourmodel(KM): 1.0}}\space\space \space\space \space \space\space\space \space\space \space \space\space\space \space\space \space \space\space\space \space\space \textcolor{blue}{\textbf{Reward Given by \ourmodel(OT): 0.1667}}
\\ \hline
\vspace{-3pt}
{\textbf{Turn 6}:}\\ <think> Okay, so Lily Chen is located in Santa Barbara. Now, the next step is to ...... </think>\\
<tool\_call> {``name'': ``building\_material\_analyzer'', ``arguments'': {``location'': ``Santa Barbara''}} </tool\_call>\\
<tool\_response> {``result'': ``Analyzing building materials in Santa Barbara for modern period. Building type: all types. Environmental impact not considered. Availability considered. Technological advancements not considered. Cultural influences not considered. Economic factors not considered. Common building material: Stone.''} </tool\_response>\\
\textcolor{DarkGreen}{[CORRECT]} \space\space\space\space\space\space \space\space \space \space\space\space \space\space \space \space\space\space \space\space \textcolor{purple}{\textbf{Reward Given by \ourmodel(KM): 1.0}}\space\space \space\space \space \space\space\space \space\space \space \space\space\space \space\space \space \space\space\space \space\space \textcolor{blue}{\textbf{Reward Given by \ourmodel(OT): 0.1667}}
\\ \hline
\vspace{-3pt}
{\textbf{Turn 7}:}\\ <think> The common building material in Santa Barbara, where the architect Lily Chen (inspired by Hearst Castle) is located, is stone. </think>\\
<answer> Stone. </answer>\\
\textcolor{DarkGreen}{[CORRECT]} \space\space\space\space\space\space \space\space \space \space\space\space \space\space \space \space\space\space \space\space \space\space \space \space\space\space \space\space \space\space \space \space\space\space \space\space \space\space \space \space\space\space \space\space \space\space \space \space\space\space \space\space \space\space \space \space\space\space \space\space \textcolor{DarkGreen}{\textbf{Outcome Reward: 1.0}}\\
\bottomrule
\end{tabular}%
\caption{A case study of turn-level reward given by \ourmodel (KM) and \ourmodel (OT).
}
\label{appendix:tab_case_study}
% \vspace{-.4cm}
\end{table*}

\begin{figure}[t]
\begin{tcolorbox}[title=Prompt Template for Training]
\small
\textbf{System Prompt:} 
\\\# Tools\\
You may call one or more functions to assist with the user query.\\
You are provided with function signatures within <tools></tools> XML tags: \\
<tools> \\
\{Tool List\}\\
</tools>\\
For each function call, return a json object with the function name and arguments within <tool\_call></tool\_call> XML tags:\\
<tool\_call>\\
\{``name'': <function-name>, ``arguments'': <args-json-object>\}\\
</tool\_call>\\
\textbf{User Prompt:} \\
Please call the given tools to answer the question. Please note that all your information must be obtained by calling tools and not by answering the question directly.\\
If the call fails, you need to try to correct it and continue until you arrive at an answer. \\
Only output the final answer (in words, numbers, or phrases) inside the <answer></answer> tag, without any explanations or extra information.\\
Question: \{question\}\\
\textbf{Output:} 
\end{tcolorbox}
\vspace{-0.2cm}
\caption{Prompt template used in our experiments.}
\label{appendix:fig_prompt}
\vspace{-0.6cm}
\end{figure}

\section{Prompt}
\label{sec:appendix_prompt}
In this section, we present the prompts used in our experiments. 
Specifically, we follow the template provided in FTRL~\cite{ye2025feedback}, the prompt template for training is shown in Figure~\ref{appendix:fig_prompt}.

\end{document}